# Ontology-based and User-focused Automatic Text Summarization (OATS): Using COVID-19 Risk Factors as an Example

**Po-Hsu Allen Chen, Amy Leibrand, Jordan Vasko, Mitch Gauthier**

Battelle Memorial Institute

{chenp, leibranda, vasko, gauthier}@battelle.org

## Abstract

This paper proposes a novel Ontology-based and user-focused Automatic Text Summarization (OATS) system, in the setting where the goal is to automatically generate text summarization from unstructured text by extracting sentences containing the information that aligns to the user's focus. OATS consists of two modules: ontology-based topic identification and user-focused text summarization; it first utilizes an ontology-based approach to identify relevant documents to user's interest, and then takes advantage of the answers extracted from a question answering model using "questions" specified from users for the generation of text summarization. To support the fight against the COVID-19 pandemic, we used COVID-19 risk factors as an example to demonstrate the proposed OATS system with the aim of helping the medical community accurately identify relevant scientific literature and efficiently review the information that addresses risk factors related to COVID-19.

## 1 Introduction

The quantity of scientific literature has dramatically increased (8-9% each year) over the last several decades (Landhuis, 2016). For example, since the COVID-19 pandemic outbreak was first identified in Wuhan, China, in December 2019 (WHO, 2020), more than 34,000 COVID-19 related scientific articles have been published in PubMed® in merely six months which makes it challenging for domain experts to keep up with newly published data. Automatic text summarization (ATS) is one of the important applications of Natural Language Processing (NLP). It is the process of automatically generating a shorter version of information from the source text either by extracting the most important sentences available in the text without alteration (extractive text summarization) or creating a summary of the original text in fewer words (abstractive text summarization). A good automatic text summarization not only saves researchers' time in reading through an entire corpus of articles but also pinpoints the answers to researchers' questions.

A common issue in an ATS system is that the generated text summarization is too general without providing specific information relevant to the user's focus to help them efficiently scan thousands of articles. Because of this, users may skip articles and lose important information, or waste time reading irrelevant articles. To address this issue, one approach to selecting user-relevant articles is to rely on the manually annotated knowledge from domain-specific documentations. For example, Schwemmer et al. (2019) manually extracted "gold standard" concept graphs from around 2,000 clinical quality measure (CQM) descriptions to guide the selection of relevant measure concepts from biomedical literature. Arumae & Liu (2019) developed a supervised deep learning framework utilizing human annotated question-answer pairs from abstracts to form a text summarization. However, those approaches are expensive and time-consuming (e.g., domain experts are required to manually review thousands of documentations) and limited (e.g., it is difficult to extend to a new domain).

In this work, we introduce a novel cognitive system, ontology-based and user-focused automatic text summarization (OATS), which consists of two modules: 1.) identification of relevant documents for a specific topic guided by an ontology with specified concepts and relations; 2.) generation of user-focused extractive text summarization from each relevant document utilizing "questions" specified from users. The ability to extract these summary sentences from unstructured text without manually annotated knowledge makes this tool unique among other ontology-based ATS systems; more importantly OATS can be applied to automatically summarize the scientific literature in gap areas in which a significant amount of evidential support exists but without existing domain documentations (e.g., COVID-19).

Since the first suspected case of COVID-19 was discovered in Wuhan, China in December 2019 (WHO, 2020), more than 60 million cases have been confirmed worldwide and the number continues to rise (Johns Hopkins University, 2020). The hallmark clinical feature of COVID-19 infections in symptomatic individuals is respiratory distress, but research shows that the virus can also impact the liver, kidney, heart, and other organ systems (Bilbul et al., 2020; Zaim et al., 2020). Furthermore, there is increasing evidence that COVID-19 infection severity and related mortality is influenced by independent risk factors, such as underlying diseases, history of disease, risky lifestyle or behaviors, gender, and age (Hamer et al., 2020; Parohan et al., 2020). Meta-analyses suggest that hypertension, diabetes, and pulmonary disease—particularly chronic obstructive pulmonary disease (COPD)—are among the most influential risk factors impacting the outcome of COVID-19 infected individuals, most notably increased severity, complications, and mortality (Parohan et al., 2020; B. Wang et al., 2020; Zheng et al., 2020). This is concerning given that these risk factors are prevalent throughout the world: in 2014, 31% of the world's population had diabetes and/or hypertension, and, in 2015, COPD was the cause of approximately 5% of all global deaths (WHO, 2014; WHO, 2017).

Uncertainty and lack of data about COVID-19 have led to the health care community effectively taking a "learn as we go" approach, in which some treatments have been advocated based on little to no evidence of efficacy (Rubin et al., 2020; Triggle et al., 2020). The extreme need for robust data to support quality health care and curb rapid spread has led to urgency within the international research community to close the knowledge gap (Osuchowski et al., 2020). Improving the understanding of the link between risk factors and COVID-19 is especially needed by clinicians to facilitate early and appropriate medical intervention and infection prevention in high risk individuals.

In support of the ongoing fight against the COVID-19, the Allen Institute for AI and other leading research groups have established the COVID-19 Open Research Dataset Challenge (CORD-19) (Wang et al., 2020), a free resource of scientific literature on COVID-19 and related diseases, to facilitate the development of NLP and other AI techniques to generate new insights. Seventeen tasks[1] (including a task to understand the impact of the risk factors of COVID-19) drawn from World Health Organization's R&D Blueprint for COVID-19 were called to action to help the medical community keep pace with the rapid acceleration in COVID-19 literature. In this paper, we use a subset of COVID-19 risk factors as an example to demonstrate the performance of the proposed automatic text summarization system.

Section 2 is a review of related works. Section 3 introduces the proposed system architecture, illustrated using the CORD-19 dataset with a focus on risk factors of COVID-19 in Section 4. Section 5 provides a summary.

## 2 Related Works

### 2.1 Ontology

One of the most popular knowledge representation approaches in NLP research is the ontology (Cambria & White, 2014), in which the ontology is often used to standardize the essential features of a topic into a concept graph representation with defined relationships between concepts. The NLP models identify and extract the concepts and relations from unstructured text, and these can then be mapped back to various concept graph representations. For example, Schwemmer et al. (2019) introduced the CQM ontology using concept graphs to represent the knowledge of CQMs. They further developed a matching algorithm to determine the relevancy of each document to a CQM using "gold standard graphs" that were manually annotated from CQM descriptions according to the CQM ontology. The same spirit is adapted in this work—we develop a COVID-19 risk factor ontology (see Subsection 3.1 for details) and identify the risk factor of COVID-19 from unstructured articles.

---

[1] https://www.kaggle.com/allen-institute-for-ai/CORD-19-research-challenge/tasks

## 2.2 NLP Models for Concept Extraction

The concepts and relations guided by the ontology can be extracted from unstructured text through processes known as Name Entity Recognition (NER) and Semantic Relation (SR) detection, respectively. These NER and SR models are often based on approaches such as a Bidirectional Long Short-Term Memory Conditional Random Field (Bi-LSTM CRF) (Huang et al., 2015), which is a deep learning method involving a type of neural network known as a long short-term memory (LSTM) layer. LSTMs have become a popular method for machine-learning tasks involving sequential data (such as sequences of words that make up sentences in text). The LSTM can also process sentences in a bi-directional manner (a Bi-LSTM), meaning that a sentence is processed both forward and backward to ensure the network obtains a more complete view of the content. Lastly, the Bi-LSTM CRF model takes the output of the Bi-LSTM and uses it as input to a conditional random field model (CRF), which have demonstrated outstanding performance for NER and SR tasks (Huang et al., 2015).

To train NER and SR models using modern deep learning-based NLP techniques (e.g., Bi-LSTM CRF), one of the biggest challenges is the shortage of training data. Because NLP is a diversified field with many distinct tasks, most task-specific datasets contain only a few thousand or a few hundred thousand human-labeled training examples. Those deep learning models see benefits as the amount of data used grows, improving when trained on millions, or billions, of annotated training examples. To minimize the challenges of the small dataset, a common approach is to utilize a pre-trained word embeddings model, which captures semantic relations and syntactic similarities between words by mapping words to low-dimensional vectors with real numbers, see Chiu et al., (2016) for an example. Word2vec (Mikolov et al., 2013) is one of the most prevalent models used to produce word embeddings through the use of neural networks. The vector representation of words allows us to capture the dependence of two words using Cosine Similarity.

## 2.3 NLP Models for Text Summarization

The vast amount of text that we encounter via articles and other sources in scientific literature can become cumbersome and, as previously stated, ATS limits researchers' need for full manual review. The general idea of extractive text summarization is to locate the most importance sentences to form a summarization. For example, Christian et al. (2016) utilizes the Term Frequency-Inverse Document Frequency (TF-IDF), a scoring metric that measures the importance of a word based on the frequency of that word in document while downplaying the score of a word in a document if it is common across the entire corpus (Salton & Buckley, 1988), to score each sentence based on the sum of the TF-IDF values for the nouns and verbs in the sentence. In short, words that are frequent in a specific document and infrequent in others will have higher scores and be categorized as keywords. Sentences with keywords hold the most pertinent information in a document, so other sentences can be filtered out to generate an accurate summarization. Although different frequency based ATS algorithms use different approaches to weight the words and sentences, a common issue of those algorithms is that the generated text summarization was constructed by the most important sentences from text but the topic of those sentences may not align with users' focuses.

## 2.4 BERT-based NLP Models

BERT (Devlin et al., 2018), a pre-trained deep bi-directional transformer model, has been used to boost the performance for all types of tasks in NLP. BioBERT (Lee et al., 2019) is also a pre-trained model, which is based on BERT and subsequently trained on biomedical domain corpora (e.g., PubMed® abstracts and PubMed Central® full-text articles). BioBERT outperforms BERT in a variety of biomedical text mining tasks because it provides a more representative word distribution for the biomedical domain. A BERT-type model can be applied with two easy steps—Pretraining and Fine-tuning. First, the network is "pretrained" to gain general knowledge of natural text. This is done by a self-supervised method in which intermittent words in sentences are randomly masked, and BERT simultaneously learns how to predict the masked word and the next sentence. Second, additional layers can be added to the end of the pretrained transformers and fine-tuned for a variety of tasks by training on labeled data. For example, Liu (2019) fine-tuned BERT-type models for text summarization by treating the extractive text summarization as a binary sentence classification problem, i.e., to predict if each sentence should be included in the summary. However, such an approach does not solve the issue as mentioned in frequency-based text summarization algorithms that the generated text summarizations would be too general without answering user's questions.

### 2.5 NLP Models for Question Answering

In addition, the pre-trained BERT model can also be fine-tuned into a question answering (QA) model—a model that can pull answers verbatim from text to respond to user-specified questions, see Zhang & Xu (2019) for an example. A general approach to training a BERT-based QA model is to add two separate classifiers to the end of the transformer model, i.e., one softmax activation layer that creates a probability distribution for the first word of the answer and another that creates a distribution for the end of the answer. Then, the model is trained to extract answers as the span of text between the maxima of the start and end distributions to match the target answers with high accuracy.

One prevalent dataset that is used to fine-tune BERT model for question answering is the Stanford Question Answering Dataset (SQuAD 2.0) (Rajpurkar et al., 2018). This is an open-source dataset consisting of 150,000 questions and thousands of passages containing answers. This dataset is differentiated from its predecessors in that 50,000 of the questions are unanswerable from the text. These unanswerable questions were crowdsourced and written to be very similar to the other questions. Therefore, this dataset enables models to not only learn to locate answers, but also to detect if none are present in the text.

### 2.6 Question Answering and Text Summarization

There is a natural connection between question answering and text summarization algorithms, both of which aim to extract important phrases or sentences. Examples of applying the question answering technique to guide the generation of summarization include those of Shi et al. (2007) who developed a system utilizing concept-level characteristics from a domain-specific ontology to summarize multiple biomedical documents to a question; Arumae & Liu (2019) who developed a supervised deep learning framework utilizing human annotated question-answer pairs from documents (abstracts) to form an extractive summarization; and Su et al. (2020) who re-ranked paragraphs based on the extracted snippets from a question answering model to generate multi-document summarizations. In addition, Chen et al. (2018) used a semantic approach for the evaluation of generated text summarization based on question answering results. All of the above approaches focused on the generation of summaries from multiple documents simultaneously; the question answering technique was used to guide the selection of sentences or paragraphs for general topics. In this work, the goal is to develop a novel text summarization tool that is capable of automatically generating extractive summaries for each relevant document separately that provide the information that aligns with the users' specific focus.

### 2.7 CORD-19 Related Research

With the release of CORD-19, multiple systems have been built to enable the exploration of valuable information related to COVID-19 by researchers and the public. For example, the host of CORD-19 created a page[2] to assemble the most useful contributions from the Kaggle community, including an interactive dashboard of summary tables for COVID-19 risk factors that were extracted by machine learning algorithms with a human curation. However, so far this contribution only covers about 3.6% of the studies published since February 2020. Other examples utilizing CORD-19 include deepMine (Joshi et al., 2020), CAiRE-COVID (Su et al., 2020), and ATS of COVID-19 (Kieuvongngam et al., 2020).

---
[2] https://www.kaggle.com/covid-19-contributions

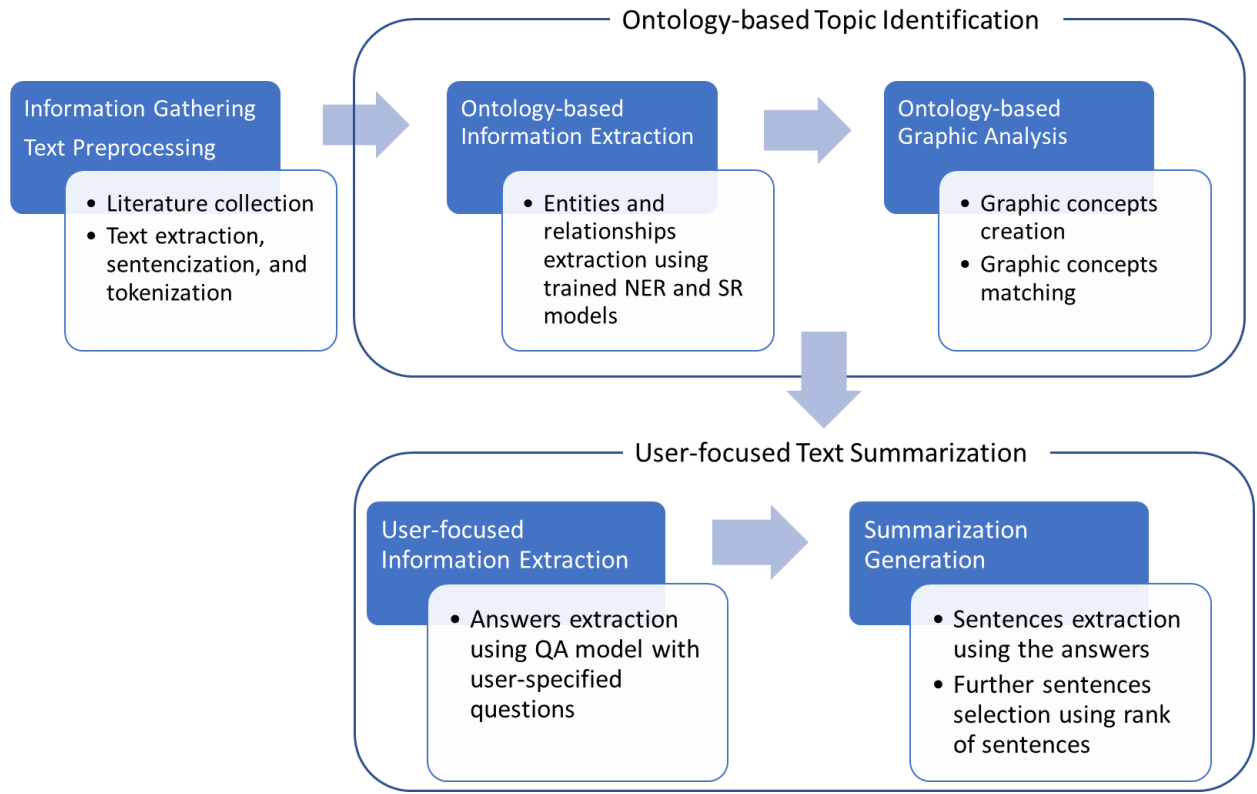

Figure 1: OATS System Architecture Overview

## 3 The System Architecture

This section illustrates the architecture of the proposed text summarization system, OATS (Figure 1), which consists of two major modules: ontology-based topic identification and user-focused text summarization.

### 3.1 Ontology-based Topic Identification

The objective of the ontology-based topic identification is to identify articles relevant to a specific topic. For example, medical experts would be interested in certain risk factors of COVID-19, e.g., diabetes, hypertension, etc. An ontology was first constructed to describe high-level knowledge about COVID-19 risk factors. This ontology guides the two major components in this module, which are 1.) NLP models to extract concepts and relations from text according to the ontology; and 2.) a graphic analytic approach to deduce the concepts within documents paired with a matching algorithm to determine the relevancy. In this paper, we will use the COVID-19 risk factors as an example to demonstrate the details of the proposed system (OATS), though the system is not limited to COVID-19 and can easily be applied to other research fields.

The goal of the COVID-19 Risk Factor ontology is to standardize features of relevant articles into a set of abstract concepts with defined relationships between them. This allows the components of relevant articles to be systematically represented, thus enabling NLP tools to identify and extract concepts and relations in a structured format that can be used for semantic reasoning and analysis. The COVID-19 Risk Factor ontology is a simplified version of a novel CQM ontology developed, in part, by two authors of this paper (Schwemmer et al., 2019). The CQM ontology focused on health care quality; however, its overarching concepts provide a broad representation of health states appropriate for capturing concepts and relations associated with COVID-19 risk factors.

| Concept | Definition | Examples |
|---|---|---|
| Population | Population and related attributes | Patients, Adults, Females |
| Health Status | Signs or symptoms, disorder, disease, complication, functional status, advanced illness | Diabetes, Hypertension, Obesity, Infection, Labored breathing |
| **Relation** | **Definition** | **(Domain, Range)** |
| IsMadeUpOf | Represents how objects combine to form composite objects | (Population, Health Status) |

Table 1: COVID-19 Risk Factor Ontology Abstract Concepts and Relations

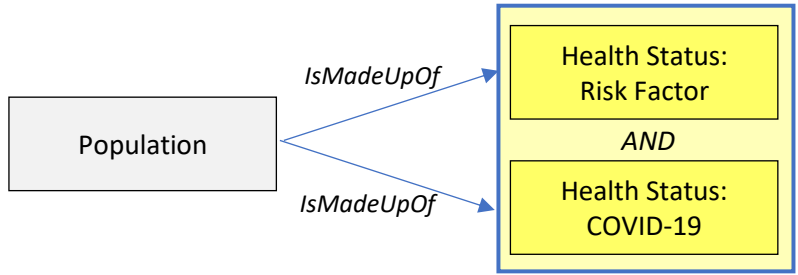

Figure 2: Graph Visualization of COVID-19 Risk Factor Ontology Concepts and Relations

| Entity (NER) | Number | Precision | Recall |
|---|---|---|---|
| Health_Status | 1,794 | 70% | 83% |
| Population | 4,062 | 73% | 79% |
| **Relation (SR)** | **Number** | **Precision** | **Recall** |
| IsMadeUpOf | 3,101 | 79% | 86% |

Table 2: Test dataset performance of NER and SR models

The CQM ontology includes a "Change Concept" component, defined as a health care behavior intended to change the progression of a health state. Although the authors recognize the value of this concept, the newness of COVID-19 presents a challenge in that the NER and SR models used for concept and relation extraction in the proposed system were trained existing NLP tools currently do not recognize "COVID-19" or its associated synonyms (i.e., the terms did not exist when the NLP models were trained), and therefore the specific relation between the two concepts cannot be efficiently recognized by the models. This complexity necessitates a simplified version of the ontology, which excludes "Change Concept" as a component and only includes a "Population" that is made up of two separate "Health Statuses": a risk factor and COVID-19. Each of these relationships between the population and a health status is considered a "triple." The abstract concepts and relations of the simplified COVID-19 Risk Factor ontology are presented in Table 1. The ontology visualized as a graph is presented in Figure 2.

Two different methods were used to extract each of the triples present in the COVID-19 Risk Factor ontology. First, "Population IsMadeUpOf Risk Factor" triple was extracted using Battelle's

NLP models. Battelle has built the NER and SR models which extract entities and relations based on the CQM ontology. Both the NER and SR models utilized a pre-trained word embeddings model that was trained on all PubMed® abstracts and PubMed Central® full text articles (Chiu et al., 2016), because it provided a good general biomedical domain word representation to help close the gap in the training data shortage and improve the ability of the NER and SR models to learn from annotated data, and therefore. This pretrained model was then fine-tuned on a set of 183 biomedical documents published prior to 2020 (74 full PubMed Central® articles and 109 PubMed® abstracts) that were manually annotated based on the CQM ontology, where 80% of data were used in training and 20% of data were for testing. The NER and SR models were both based on the Bi-LSTM CRF architecture. Table 2 provides the summary of the number of "Population" and "Health Status" concepts and the corresponding semantic relation instances manually annotated in the biomedical articles, and precision (the fraction of identified concepts that are correct) and recall (the fraction of the correct concepts that are successfully identified) results obtained by Battelle's NLP models for the test dataset.

Second, as mentioned, COVID-19 did not exist when Battelle's NLP models were trained. Manual annotation of scientific articles for creating a training dataset is a time consuming and expensive process. To extract the other triple ("Population IsMadeUpOf COVID-19") in the COVID-19 Risk Factor ontology from unstructured text without spending additional time and budget, an acceptable but imperfect solution is developed. A list of synonyms of COVID-19 was created and used to extract the COVID-19 related entities from text based on whether any terms in a sentence matched the synonyms. Then, if the sentence with a COVID-19 entity extracted also has the "Population" entity extracted from the trained NER model, the "IsMadeUpOf" relation is automatically added for connecting the two entities to form a triple.

Next, we use the igraph package (Csardi & Nepusz, 2006) in the R programming language to create the graph concepts from extracted triples guided by the COVID-19 Risk Factor ontology to represent the knowledge in each document. The relevancy of a document to a risk factor of COVID-19 (e.g., hypertension) is determined using a word2vec model that was trained on 10 years of PubMed® abstracts. If the distance (1 minus Cosine Similarity) of the "Health Status Risk Factor" from the document is less than the threshold for that particular risk factor, then the document is determined as relevant document to that risk factor.

**3.2 User-focused Text Summarization**

When the topic of a document is determined through the ontology-based topic identification module and is relevant to a user's interest, the document will go through the user-focused text summarization module to generate a text summarization that provides the information relevant to the user's questions. The two major components in this module are 1.) a QA model that can extract answers verbatim from unstructured text using user-specified questions; and 2.) a summarization generation process that selected sentences using extracted answers to form a summarization.

In this module, we first fine-tuned the BioBERT model (BioBERT-Base V1.1[3]) with the full SQuAD 2.0 dataset[4] for training a QA model. This BioBERT-based QA model works by calculating a probability distribution for the beginning and the end terms of an answer given a specific question. The most probable beginning and end are then used to extract the answer verbatim from the text.

Since the extracted answers from the QA model may appear in multiple sentences in the text, a sentence scoring method is developed to improve the performance of the summarization generation. Each sentence is tokenized and a score is assigned as the sum of Term-Frequency values for non-common words in that sentence. With common words being filtered out, there is no need for IDF normalization. For answers that have multiple corresponding sentences, only the sentence with highest score will be selected. The final step is to further down-select the sentences returned by the QA

---

[3] https://github.com/dmis-lab/biobert

[4] https://rajpurkar.github.io/SQuAD-explorer/

model using sentence scores to generate text summarization.

Note that this module is highly customizable. First, the OATS system requires users to define questions based on their interests, then users may determine the order of questions (sentences) in the generated text summarization. Different applications with different question-answering pairs would follow different logical orders for generating a summarization. The following section demonstrates the use of COVID-19 risk factors as an example of question selection and question order determination.

## 4 Experimental Results

The CORD-19 dataset is actively updated by Allen Institute for AI. The full dataset was downloaded (updated on June 22[nd], 2020), in which there were more than 130,000 COVID-19-related scientific publications. The dataset includes a set of target tables providing the COVID-19-related findings by the COVID-19 Kaggle community contributions. Particularly, for risk factors of COVID-19, findings such as Study Population, Sample Subjects, Study Type, Severe Label, and Fatality Label corresponding to 28 risk factors in 217 articles were provided, from which 104 full text articles were freely available from the CORD-19 dataset. In this section, the performance of the proposed text summarization system, OATS, is examined using these 104 articles in terms of ability to identify the risk factor(s) and the accuracy of text summarization.

### 4.1 Topic Identification on CORD-19 dataset

Four influential risk factors (hypertension, diabetes, obesity, and COPD) impacting the outcome of COVID-19-infected individuals were selected. We manually reviewed 104 articles and determined whether each article was relevant (i.e., the article was related to one of the influential risk factors). Table 3 shows the performance of the topic identification module in OATS. In general, OATS has high precision scores and acceptable recall scores on all four COVID-19 risk factors, which suggests that using the ontology-based method to determine the topic of an article is appropriate and the chance to provide a false positive result is low.

| Risk Factor | Precision | Recall |
|---|---|---|
| Hypertension | 0.93 | 0.84 |
| Diabetes | 0.92 | 0.76 |
| Obesity | 1 | 0.71 |
| COPD | 0.81 | 0.6 |

Table 3: Performance of the Topic Identification Module in OATS

The system has a slightly poor performance with the risk factor COPD. One potential reason may be that the term "COPD" is an acronym, which is more difficult for the trained word2vec model to correctly measure distance with other terms. For example, if only "COPD" and its full name "chronic obstructive pulmonary disease" were used to find matching terms, the recall score would be low; however, if we separated COPD into four words and measured the distance from each word to the candidate term, the precision score would be low because the term "chronic" is commonplace and thus many terms exhibit a relatively small distance from it. Ideas for improvement includes adding an algorithm to extract an acronym's corresponding full name or task-specific training to improve the word2vec model.

All full text articles in the CORD-19 dataset were processed using OATS to identify articles relevant to the four risk factors of COVID-19 investigated in this paper, in which 9% of CORD-19 articles were found to be relevant to hypertension, 6.5% relevant to diabetes, 1.2% were relevant to obesity, and 4.4% were relevant to COPD[5].

### 4.2 Text Summarization on CORD-19 Dataset

Relevant articles identified from the ontology-based topic identification module will then be processed through the user-focused text summarization module. In this section, one risk factor of COVID-19, hypertension, will be used to demonstrate setting up "Questions" for extracting answers from the trained QA model, and the process of generating text summarization.

---

[5] The full lists of articles relevant to the four risk factors of COVID-19 can be downloaded from https://battelle.box.com/s/qhapve1j6utjgcb5m0safkghcwvqw4ti.

| Questions | True answer | Extracted Answer |
|---|---|---|
| Q1 Are patients with hypertension? | Patients with hypertension were observed in 16.8% of the 573 patients with abnormal CT imaging/pneumonia | Patients with at least one coexisting underlying conditions and patients with hypertension were observed in 28.8% and 16.8% |
| Q2 Which hospital is studied? | Multiple hospitals in Zhejiang, China | Zhejiang China |
| Q3 What is the date of the study? | January 17 to February 8, 2020 | January 17 to February 8 |
| Q4 Is this a prospective observational study, retrospective observational study, or systematic study? | Retrospective observational | retrospective |
| Q5 How many patients are in this study? | 645 | 645 |
| Q6 How many studies are in this article? | [No text extracted] (only one study) | COVID-19 |
| Q7 Is there a hypertension odds ratio for fatality patients? | No text extracted (fatality not studied) | significantly higher than the non-pneumonia patients all P < 0.05 |
| Q8 Is there a hypertension odds ratio for severe patients? | Patients with… hypertension were observed in… 16.8% of the 573 patients … which was significantly higher than the non-pneumonia patients (all P < 0.05). | significantly higher than the non-pneumonia patients all P < 0.05 |

Table 4: Examples of Specified Questions and Extracted Answers

After creating summary tables that address risk factors related to COVID-19[6] based on the CORD-19 Challenge, eight questions (see Table 4 for details) were used in the QA model for extraction of "answers" from text. These questions followed logic to extract risk factor-related information, as well as study location, date(s) and type. Follow-up questions aimed to obtain additional study details, i.e., cohort size or fatality and/or severity odds ratio data.

The title of the example used in Table 4 is "Epidemiological, clinical characteristics of cases of SARS-CoV-2 infection with abnormal imaging findings" (Zhang et al., 2020). Table 4 presents the *True answers* curated by manual (human) extraction and the *Extracted answers* from the trained QA model. Extracted answers were then used to select sentences based on the sentence scores obtained from the sentence ranking

---

[6] https://www.kaggle.com/allen-institute-for-ai/CORD-19-research-challenge/tasks?taskId=888

> **Patients with at least one coexisting underlying conditions and patients with hypertension were observed in 28.8% and 16.8%** of the 573 patients respectively, which was significantly higher than the non-pneumonia patients all P < 0.05. For this retrospective study, 645 patients confirmed with SARS-CoV-2 infection between January 17 and February 8, 2020 underwent a CT examination or X-ray, in **Zhejiang, China**. Patients confirmed with SARS-CoV-2 infection in Zhejiang province from **January 17 to February 8** who had undergone CT or X-ray were enrolled. In our **retrospective** study, we evaluated and compared the epidemiological clinical features and laboratory data of those with abnormal imaging findings. The imaging findings of SARS-CoV-2 pneumonia are similar to acute respiratory syndrome SARS and Middle East respiratory syndrome MERS which are characterized as pulmonary ground-glass opacities and consolidation (Das et al. 2016). 139 (21.5%) patients of the total **645** patients had one affected lobe, 204 (31.6%) patients had two affected lobes, 136 (21.1%) patients had three lobes affected, 66 (10.2%) had four affected lobes, and (28 4.4%) patients had five affected lobes. Finally, according to the admission data risk factors for severe critical type of **COVID- 19** were identified; however, we still lack a prediction model for disease progression. In conclusion there are certain characteristics of the chest imaging of COVID-19 patients we reported the differences in specific epidemiological and clinical features between patients with abnormal or normal imaging including fever cough and sputum production and relatively poor laboratory results.

Figure 3: Example of Generated Text Summarization with bold font indicating the extracted answers in the sentences

algorithm. The generated summarization (shown in Figure 3—bold font indicates extracted answers) provides information about percentage of COVID-19 patients with hypertension, followed by study dates (e.g., January 17 to February 8, 2020), location (e.g., Zhejiang China), and so on. This text summarization provides medical experts a quick indication of the worth of the full article text as it relates to their interest in hypertension.

To better understand the performance from the user-focused text summarization module, text extracted from nineteen articles[7] within the set of 104 articles was further evaluated using a QA model for assessing accuracy. Two raters independently scored answers (i.e., extracted text) to eight questions as well as the overall accuracy of the text summarization. Scores that disagreed were reviewed jointly by the raters for discussion and consensus. Criteria for scoring varied by question. Questions and scoring criteria for each are presented in Table 5. Note that for Q7 and Q8, two versions of each question were used to assess how wording impacted the results.

Results of the Question Answering evaluation are presented in Figure 4, below. Note that Q7 and Q8 scores improved by adjusting the wording of questions to be more specific to the risk factor of interest, and the improved versions (Q7_B and Q8_B) are used to generate the extractive summary. In general, the QA model responses and the derived text summaries were encouraging considering the difficulty of the task and the lack of domain-specific fine-tuning of the model. The extracted answers pertaining to methodology information (Q1 through Q6) had over 50% accuracy (which is a competitive rate as shown in Lee et al., (2019)), and, in many cases, demonstrated that the model can distinguish between highly similar numerical details, such as the number of patients enrolled versus of the number of patients screened.

---

[7] The first 20 articles were selected. One article reviewed was not relevant to hypertension, and thus was skipped during score comparison.

| Question | Type* | Scoring Criteria | |
|---|---|---|---|
| | | 1 | 0 |
| Q1 Are patients with hypertension? | Qual. | Useful information about hypertension with context | No useful information about hypertension |
| | | | No text extracted |
| Q2 Which hospital is studied? | Qual. | Correct name of hospital where study occurred | Extracted text did not provide name of hospital where study occurred |
| | | No text extracted because study was not performed in a hospital | No text extracted, but study was performed in a hospital |
| Q3 What is the date of the study? | Quan. | Start, end or range of study dates | Incorrect date or no text extracted |
| Q4 Is this a prospective observational study, retrospective observational study, or systematic study? | Qual. | Correct type of study extracted | Incorrect study or no text extracted |
| Q5 How many patients are in this study? | Quan. | Correct number of patients studied | Incorrect number of patients or no text extracted |
| Q6 How many studies are in this article? | Quan. | Correct number of studies | Incorrect number of studies |
| | | No text extracted because only one study was performed (not explicitly stated in article) | No text extracted, but multiple studies were performed |
| Q7_A Is there an odds ratio for fatality patients? | Quan. | Correct odds ratio for mortality in hypertensive patients | Incorrect odds ratio for mortality in hypertensive patients |
| Q7_B Is there a hypertension odds ratio for fatality patients? | | Data related to risk associated with mortality in hypertensive patients | Odds ratio for condition other than mortality in hypertensive patients |
| | | | No text extracted |
| Q8_A Is there an odds ratio for severe patients? | Quan. | Correct odds ratio for severity in hypertensive patients | Incorrect odds ratio for severity in hypertensive patients |
| Q8_B Is there a hypertension odds ratio for severe patients? | | Data related to risk associated with severity in hypertensive patients | Odds ratio for condition other than severity in hypertensive patients |
| | | | No text extracted |
| Text summarization paragraph | Qual. | Summary provided reasonable details about study design and relevance to hypertension (score = 1) | Summary did not provide reasonable details about study design and relevance to hypertension (score = 0) |

*Quan. denotes Quantitative, and Qual. Denotes Qualitative

Table 5: Questions and Scoring Criteria for QA Model

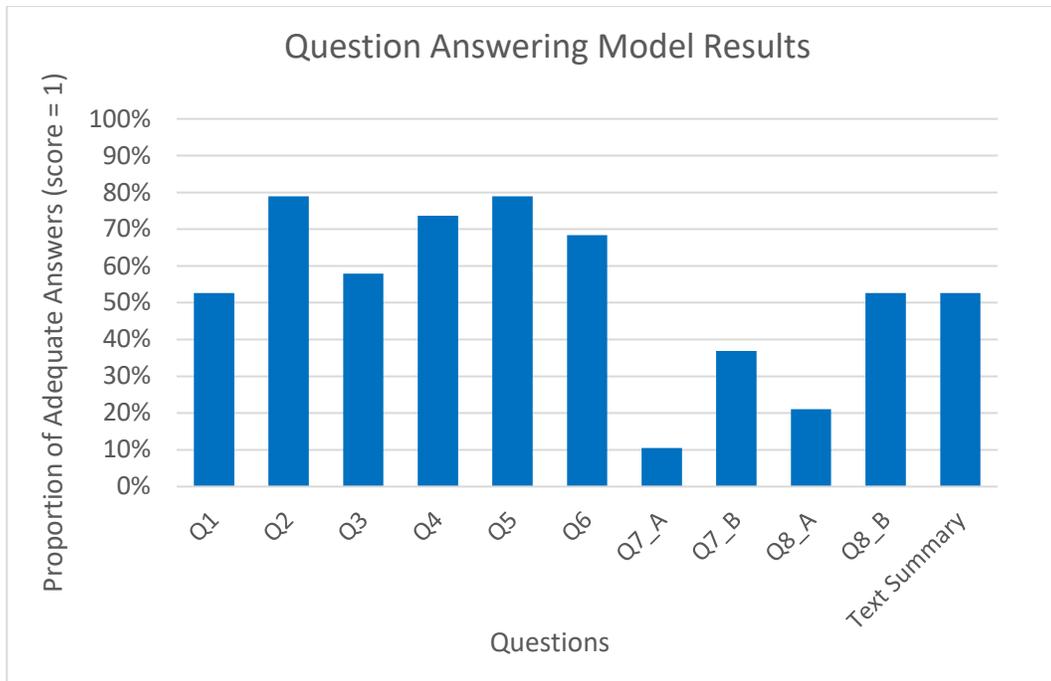

Figure 4: QA Evaluation Results

Further, the QA model is able to recognize when the answer is not present in the paper and resultingly returns no text (e.g., no specific hospital is reported for Q2 when the study was a meta-analyses). The relatively high performance on these questions is likely due to the articles typically stating the corresponding answers using standard language, which likely facilitates the matching of the most relevant text to the wording of the question.

A limitation is that the QA extraction accuracy can be sensitive to the wording of the question, especially for information that is presented slightly differently across documents. For example, when the word "hypertension" was included in Q7_B and Q8_B, the QA model extracted the correct odds ratio for risk from hypertension at a higher rate when stated in a study, and was more likely to return no information when no explicit odds ratio was stated in the available text. An additional limitation particular to Q7 and Q8 is that odds ratios are often presented in tables, which are not included in the text available to the QA model. Similarly, preprocessing difficulties on passages containing odds ratios caused them to be dropped from the text presented to the QA model. Further, the QA model is designed to find the single extractive summary that best answers the question. Answers that are completed across multiple sentences or in multiple locations across the article therefore may be missed by the model.

However, the performance on the individual questions is not directly reflected in the quality of the overall text summary produced by the OATS model. Incorrect text extracted by the QA model can still lead to inclusion of useful sentences in the summary produced by the OATS model. For example, in one document an odds ratio related to increased risk from hypertension was not included in the preprocessed text available to the QA model, although the hypertension odds ratio was noted in the full document text (i.e., the odds ratio data was unintentionally discarded during preprocessing of text). The QA model correctly returned no text for Q7_B and Q8_B, and subsequently the extractive summary includes no explicit information on hypertension. In contrast, while the model incorrectly returns "conclusive evidence as risk factors" for Q7_A and Q8_A, this leads to the inclusion of an informative sentence in the final summary: "This study shows the conclusive evi-dence as *risk factors for severe COVID-19 patients such as patient characteristics age gender BMI comorbidities DM HTN CHD COPD* vital signs respiratory rate symptoms fever and dyspnea and laboratory findings blood routine biochemical indicators inflammatory biomarkers and coagulation..." This finding complicates the question-writing process and may lead to difficulty in fine-tuning questions for optimal summaries.

**5 Summary**

This paper proposes a novel automatic text summarization system, called OATS, which first utilizes an ontology-based approach using concept graphs to represent the knowledge from unstructured text and a matching algorithm to identify relevant documents, while also exploiting the answers from a question answering model based on "questions" specified by users to generate text summarization that aligns to the user's focus.

In general, the results are encouraging considering the difficulty of the task and the lack of domain-specific fine-tuning of the QA model. A limitation is that the QA model approach requires a standardized set of questions to access information that may be presented in non-standardized ways across documents (e.g. risk from hypertension). Furthermore, inaccessibility of the sections of the text that contain the optimal answers to questions remain can limit QA model performance. There can also be a disconnect between the performance on question answering and the quality of the overall summary, where the best sentences to include may not always be captured by the answers a standardized set of questions. Sometimes the performance on the individual questions is not directly reflected in the quality of the overall summary. Incorrect text extracted by the QA model may still result in inclusion of useful sentences in the summary. This may make it difficult to fine-tune the questions for optimal results. Regardless of these limitations, the model accuracy was on part with state-of-the-art question-answering models, and was able to generate an acceptable summarization in over half of the documents reviewed.

# Appendix

## Additional Example 1: High-performing example

**Document Title:** Neutrophi-to-lymphocyte ratio as an independent risk factor for mortality in hospitalized patients with COVID-19

| Question | True answer | Extracted Answer | Score |
| --- | --- | --- | --- |
| Q1 Are patients with hypertension? | In the multivariate adjusted models, … hypertension [was] included. | Hypertension OR = 3.94 95%CI 1.82-8.53 | 1 |
| Q2 Which hospital is studied? | Zhongnan Hospital of Wuhan University | Zhongnan Hospital of Wuhan University | 1 |
| Q3 What is the date of the study? | January 1 to February 29, 2020 | January 1 to February 29 2020 | 1 |
| Q4 Is this a prospective observational study, retrospective observational study, or systematic study? | retrospective observational/ retrospective cohort study | retrospective cohort study | 1 |
| Q5 How many patients are in this study? | 245 | 245 | 1 |
| Q6 How many studies are in this article? | [No text extracted] (only one study) | [No text extracted] | 1 |
| Q7_A Is there an odds ratio for fatality patients? | The univariate analysis concluded that… Hypertension (OR = 3.94, 95%CI, 1.82-8.53, P = 0.0005) [was] positively correlated with the risk of in-hospital death | unadjusted and adjusted odds ratio ORs | 0 |
| Q7_B Is there a hypertension odds ratio for fatality patients? | | Hypertension OR = 3.94 95%CI 1.82-8.53 P = 0.0005 | 1 |
| Q8_A Is there an odds ratio for severe patients? | No text extracted (severity not studied) | unadjusted and adjusted odds ratio ORs | 0 |
| Q8_B Is there a hypertension odds ratio for severe patients? | | [No text extracted] | 1 |
| Text summarization paragraph | The univariate analysis indicated that age (OR = 1.09, 95% CI, 1.06-1.13, P < 0.0001), BMI (OR = 1.15, 95% CI, 1.01-1.30, P = 0.0328), Hypertension (OR = 3.94, 95%CI, 1.82-8.53, P = 0.0005), Diabetes (OR = 3.30, 95% CI, 1.24-8.77, P = 0.0168), CHD (OR = 6.46, 95% CI, 2.33-17.90, P = 0.0003), Respiratory rate ≥ 30 bpm (OR = 7.43, 95% CI, 1.76-31.38, P = 0.0064), Neutrophil (OR = 1.34, 95% CI, 1.19-1.50, P < 0.0001), ALT (OR = 1.01, 95% CI, 1.00-1.02, P = 0.0094), Creatinine (OR = 1.01, 95% CI, 1.00-1.01, P = 0.0177), PT (OR = 1.30, 95% CI, 1.01-1.67, P = 0.0448), C-reactive protein and Procalcitonin values were positively correlated with the risk of in-hospital death. For this retrospective cohort study, COVID-19 patients who were admitted to Zhongnan Hospital of Wuhan University from January 1 to February 29, 2020 were consecutively included. Third, although we have adjusted for multiple potential confounders, residual and unmeasured confounding might not be fully considered. In conclusion, this retrospective cohort study performed in the Chinese population revealed that the NLR is an independent risk factor for the in-hospital mortality. Discussion This retrospective cohort study included 245 COVID-19 patients and the total in-hospital mortality was 13.47%. | | |

**Additional Example 2: Low-performing example**

**Document Title:** Prediction of the clinical outcome of COVID-19 patients using T lymphocyte subsets with 340 cases from Wuhan, China: a retrospective cohort study and a web visualization tool.

| Question | True answer | Extracted Answer | Score |
|---|---|---|---|
| Q1 Are patients with hypertension? | Discharged cases: the most common underlying disease is Hypertension, 23.9% for male and 20.0% for female<br>Death cases: the most common underlying disease is also<br>Hypertension, 64·7% for male and 38·5% for female | 23.9% for male and 20.0% for female | 1 |
| Q2 Which hospital is studied? | Wuhan Pulmonary Hospital | Wuhan Pulmonary Hospital | 1 |
| Q3 What is the date of the study? | January 31 to March 8, 2020 | [No text extracted] | 0 |
| Q4 Is this a prospective observational study, retrospective observational study, or systematic study? | Retrospective cohort study | [No text extracted] | 0 |
| Q5 How many patients are in this study? | 340: (310 discharged cases and 30 death cases) | 310 | 0 |
| Q6 How many studies are in this article? | [No text extracted]<br>(only one study) | 310 | 0 |
| Q7_A Is there an odds ratio for fatality patients? | Discharged cases: the most common underlying disease is Hypertension, 23.9% for male and 20.0% for female<br>Death cases: the most common underlying disease is also<br>Hypertension, 64·7% for male and 38·5% for female | [No text extracted] | 0 |
| Q7_B Is there a hypertension odds ratio for fatality patients? | | [No text extracted] | 0 |
| Q8_A Is there an odds ratio for severe patients? | [No text extracted]<br>(severity not studied) | [No text extracted] | 1 |
| Q8_B Is there a hypertension odds ratio for severe patients? | | [No text extracted] | 1 |
| Text summarization paragraph | With the COVID-19 becoming a pandemic all over the world we aim to share our epidemiological and clinical findings with the global community. In this retrospective cohort study, we studied 340 confirmed COVID-19 patients from Wuhan Pulmonary Hospital including 310 discharged cases and 30 death cases. Excluding four patients whose direct cause of death was not COVID-19 infection and selecting patients who had at least one T cell Subsets test available, we had a total of 340 patients in the study including 310 discharged cases and 30 death cases. We reviewed laboratory test results and chest CT examinations of these 340 patients and collected all the T lymphocyte subsets tests data. | | 0 |